\documentclass[lettersize,journal]{IEEEtran}
\usepackage{amsmath,amsfonts}
\usepackage{algorithmic}
\usepackage{algorithm}
\usepackage{array}
\usepackage[caption=false,font=normalsize,labelfont=sf,textfont=sf]{subfig}
\usepackage{textcomp}
\usepackage{stfloats}
\usepackage{url}
\usepackage{verbatim}
\usepackage{graphicx}
\usepackage{cite}

\usepackage{amssymb}
\usepackage{mathtools}
\usepackage{booktabs}
\usepackage{pifont}
\usepackage{float}
\usepackage{bbding}
\usepackage{wrapfig}
\usepackage{multirow}
\usepackage{multicol}
\usepackage[table]{xcolor}
\usepackage{tcolorbox}
\usepackage{hyperref}
\usepackage{mdframed}

\usepackage{float}
\usepackage{algorithm}
\usepackage{booktabs}
\usepackage{amsfonts}
\usepackage{graphicx}
\usepackage{amsmath}
\usepackage{arydshln}
\usepackage{multirow}
\graphicspath{{figures/}}
\usepackage{bbding}
\usepackage[table]{xcolor}
\usepackage{color}
\definecolor{mygray}{gray}{.9}

\begin{document}

\title{Branch-Tuning: Balancing Stability and Plasticity for Continual Self-Supervised Learning}
\author{Wenzhuo Liu,
		Fei Zhu,
		Cheng-Lin Liu~\IEEEmembership{Fellow,~IEEE}
	\IEEEcompsocitemizethanks{\IEEEcompsocthanksitem Wenzhuo Liu and Cheng-Lin Liu are with the University of Chinese Academy of Sciences, Beijing, P.R. China, and the State Key Laboratory of Multimodal Artificial Intelligence Systems, Institute of Automation, Chinese Academy of Sciences, 95 Zhongguancun East Road, Beijing 100190, P.R. China. \protect
    \IEEEcompsocthanksitem Fei Zhu is with the Centre for Artificial Intelligence and Robotics, Hong Kong Institute of Science and Innovation, Chinese Academy of Sciences, Hong Kong 999077, China. \protect
		\IEEEcompsocthanksitem  Email: liuwenzhuo20@mails.ucas.ac.cn, fei.zhu@cair-cas.org.hk, liucl@nlpr.ia.ac.cn}
}


\markboth{Journal of \LaTeX\ Class Files,~Vol.~14, No.~8, August~2021}%
{Shell \MakeLowercase{\textit{et al.}}: A Sample Article Using IEEEtran.cls for IEEE Journals}


\maketitle

\begin{abstract}
Self-supervised learning (SSL) has emerged as an effective paradigm for deriving general representations from vast amounts of unlabeled data. However, as real-world applications continually integrate new content, the high computational and resource demands of SSL necessitate continual learning rather than complete retraining. This poses a challenge in striking a balance between stability and plasticity when adapting to new information. 
In this paper, we employ Centered Kernel Alignment for quantitatively analyzing model stability and plasticity, revealing the critical roles of batch normalization layers for stability and convolutional layers for plasticity. 
Motivated by this, we propose Branch-tuning, an efficient and straightforward method that achieves a balance between stability and plasticity in continual SSL.
Branch-tuning consists of branch expansion and compression, and can be easily applied to various SSL methods without the need of modifying the original methods, retaining old data or models. We validate our method through incremental experiments on various benchmark datasets, demonstrating its effectiveness and practical value in real-world scenarios. We hope our work offers new insights for future continual self-supervised learning research. The code will be made publicly available.
\end{abstract}


\ifCLASSOPTIONcompsoc
\IEEEraisesectionheading{\section{Introduction}\label{sec:introduction}}
\else
\section{Introduction}
\label{sec:introduction}
\fi

\IEEEPARstart
Self-supervised learning (SSL) has attracted widespread attention from researchers in recent years, as it allows models to learn from unlabeled data without human annotations, providing a more versatile and promising alternative to supervised learning \cite{chen2020simple, he2020momentum,zbontar2021barlow,caron2020unsupervised,caron2018deep}. Although advantageous, SSL requires more resources to achieve supervised learning performance, resulting in substantial computational costs \cite{chen2020simple,he2020momentum,caron2020unsupervised}.
In real-world scenarios where data is constantly generated, such as the emergence of trending topics on social media platforms or the evolution of consumer behavior and preferences in online shopping environments \cite{luo2020alicoco,xu2019open}, it is impractical to repeatedly collect data and retrain models. Therefore, SSL models must be designed to continually adapt to new information.
\begin{figure}[t]
  \centering
  \vskip 0.05in
  \includegraphics[width=1.0\linewidth]{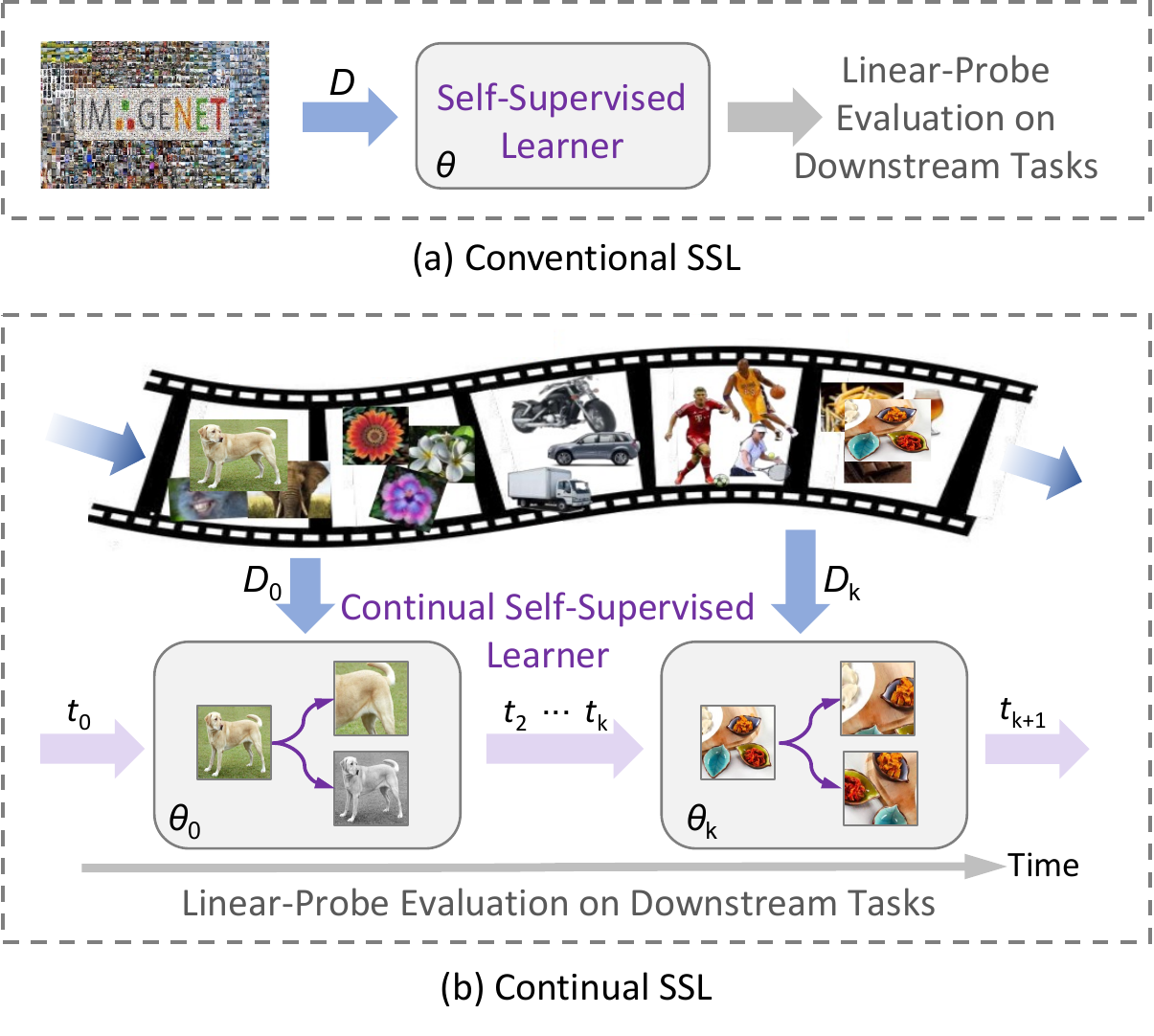}
  \vskip -0.1in
  \caption{The difference between (a) Conventional SSL and (b) Continual SSL. In real-world scenarios, non-IID and infinite data emerge continuously. Continual SSL leverages this data flow to train self-supervised models.}
  \vskip -0.1in
  \label{fig:1}
\end{figure}

To incorporate new knowledge, an intuitive way is to train on new data directly, a process known as Fine-tuning \cite{yosinski2014transferable}. However, Fine-tuning can lead to catastrophic forgetting, which degrades the model's performance on previously learned tasks \cite{9145832,french1999catastrophic,robins1995catastrophic,mccloskey1989catastrophic,zhangCvpr22ContinuSSeg}. This issue arises due to the high plasticity of neural networks, as parameters crucial for old tasks may be updated to better adapt to new data \cite{kim2023stability}. Consequently, self-supervised models face a dilemma between stability and plasticity: Fine-tuning on new data undermines stability, while not updating the model sacrifices the plasticity needed for adapting to new information.

In the context of supervised learning, continual learning, also known as incremental learning, has been proposed to address catastrophic forgetting in neural networks \cite{9724647,9181489,rebuffi2017icarl,zhu2022learning,hou2019learning}. Widely used methods include network regularization \cite{kirkpatrick2017overcoming,aljundi2018memory,li2017learning,dhar2019learning}, knowledge distillation \cite{hinton2015distilling,9899753,cui2023uncertainty}, and data replay \cite{belouadah2018deesil,castro2018end,10058177}.
Recognizing these developments, some studies \cite{cha2021co2l,purushwalkam2022challenges,tang2023practical} have explored the challenge of continual SSL and proposed solutions that mainly borrow techniques from supervised learning, such as data replay and knowledge distillation. In continual SSL, the focus lies on learning feature extractors, and the performance of state-of-the-art continual methods in supervised learning does not necessarily apply to SSL \cite{fini2022self}. Moreover, these methods inherit issues from supervised continual learning, such as needing to store old data to prevent forgetting and saving old models for knowledge distillation, which often requires designing intricate distillation techniques.

In this work, we systematically analyze the stability and plasticity of models in continual SSL, which are defined and quantified using Centered Kernel Alignment (CKA) metric \cite{cortes2012algorithms,kornblith2019similarity}.  Our experiments reveal that in convolutional networks, Batch Normalization (BN) layers are crucial for model stability, while convolutional layers are essential for model plasticity. To achieve the desired balance, the training of old and new parameters in the convolutional layers is separated. We introduce branches with new parameters, enabling learning from new data while preserving old parameters. After that, reparameterization \cite{ding2021repvgg, ding2021diverse, ding2019acnet,ding2021resrep} is employed to equivalently compress the new parameters, ensuring the model structure remains unaltered. The learning process on new data comprises branch expansion and branch compression, a technique we term as Branch-tuning. This approach can be simply and efficiently applied to various SSL methods and continual SSL methods by replacing Fine-tuning with Branch-tuning, without the need of modifying the original methods, retaining old data or old models, or performing cumbersome knowledge distillation. Experiments demonstrate the effectiveness of our method in different settings, including incremental experiments on the TinyImageNet, CIFAR-100 \cite{krizhevsky2009learning}, and ImageNet-100 \cite{deng2009imagenet} benchmark datasets, and transfer experiments on the real-world dataset RAF-DB \cite{li2017reliable}.

Our contributions are as follows:

\begin{itemize}
\item We provide a quantitative analysis of stability and plasticity for models in continual SSL. Our investigation into the roles of network layers reveals that fixed BN layers improve stability, while adaptable convolutional layers promote plasticity.
\item Based on these findings, we propose Branch-tuning to balance stability and plasticity in continual SSL. Our method eliminates the need of retaining old models or data and can be implemented without modifying existing SSL methods.

\item Our method demonstrates improvements in various incremental tasks such as class incremental learning, data incremental learning, and real-world dataset transfer learning, increasing the practicality of SSL models in real-world scenarios.
\end{itemize}
The rest of this paper is organized as follows. Section \ref{sec:Related Work} reviews the related work and defines the CIL problem in Section \ref{sec: 3}; Section \ref{sec:4} presents the quantitative analysis of the stability and plasticity of the model; Section \ref{sec:5} describes the proposed approach; Section \ref{sec:6} presents our experimental results, and Section \ref{sec:7} draws concluding remarks.


\section{Related Work}
\label{sec:Related Work}
\subsection{Self-Supervised Learning}
Self-supervised learning (SSL) utilizes unlabeled data to extract useful representations through pretext tasks such as rotation prediction \cite{gidaris2018unsupervised}, patch position determination \cite{doersch2015unsupervised}, image colorization \cite{larsson2016learning}, and inpainting \cite{pathak2016context}. These representations are then applied to supervised tasks with labeled data.
Recent advancements in SSL can be categorized into four main approaches: contrastive learning, negative-free methods, clustering-based methods, and redundancy reduction-based methods.

\textbf{\textit{Contrastive learning methods}}, including CPC \cite{oord2018representation}, AMDIM\cite{bachman2019learning}, CMC \cite{tian2020contrastive}, SimCLR \cite{chen2020simple}, and MoCo  \cite{he2020momentum}, use augmented views of the input data to create positive and negative pairs. The model is trained to produce similar representations for positive pairs and dissimilar representations for negative pairs. 

\textbf{\textit{Negative-free methods}}, notably BYOL \cite{grill2020bootstrap} and SimSiam \cite{chen2021exploring}, only use positive pairs to learn representations, avoiding the need for negative pairs. They use two Siamese networks,  where the representations of two views must match.  One view is generated by an online and predictor network and the other by a target network.

\textbf{\textit{Clustering-based methods}}, such as SwAV \cite{caron2020unsupervised}, DeepCluster v2 \cite{caron2018deep}, and DINO \cite{caron2021emerging}, utilize unsupervised clustering algorithms to align representations with their corresponding prototypes. The prototypes are learned by aggregating the feature representations within each cluster. As a result, representations are trained to resemble their prototypes, leading to enhanced generalization and robustness. 

\textbf{\textit{Redundancy reduction-based methods}}, like BarlowTwins \cite{zbontar2021barlow} and VicReg \cite{bardes2022vicreg}, reduce the correlation between different representations. BarlowTwins optimizes a cross-correlation matrix to be unit matrix, making positives similar and negatives orthogonal, while VicReg uses a combination of variance, invariance, and covariance regularizations.

Compared to supervised learning, SSL requires more resources, such as more data and larger models, to match the performance of supervised learning \cite{chen2020simple,he2020momentum}. This means that the cost of training SSL models is higher. In real-world scenarios, where data is constantly generated, models need to continuously adapt to new knowledge rather than being retrained.

\subsection{Continual Learning}
Continual learning, also known as lifelong or incremental learning, focuses on enabling models to learn and adapt to new tasks or data without forgetting previously acquired knowledge. Continual learning methods can be broadly classified into three categories: regularization-based methods, memory-based methods, and parameter isolation-based methods.

\textbf{\textit{Regularization-based methods}}, such as EWC \cite{kirkpatrick2017overcoming}, SI \cite{zenke2017continual}, MAS \cite{aljundi2018memory}, LwF \cite{li2017learning}, SSRE \cite{zhu2022self}, and PASS \cite{zhu2021prototype}, introduce regularization terms to the loss function to protect important parameters from drastic updates. These methods can prevent catastrophic forgetting and allow the model to adapt to new tasks with minimal interference. However, they may require additional hyperparameters or introduce computational overhead. 

\textbf{\textit{Memory-based methods}}, also called replay-based or exemplar-based methods, including GEM \cite{lopez2017gradient}, A-GEM \cite{chaudhry2018efficient}, iCaRL \cite{rebuffi2017icarl}, BiC \cite{wu2019large}, PODnet \cite{douillard2020podnet}, DER \cite{yan2021dynamically}, RMM  \cite{liu2021rmm}, and FeTrIL \cite{petit2023fetril}, store a small subset of previous experiences to be used during training on new tasks.
These methods enable the model to retain previously learned knowledge while updating its parameters for the new task, and have proven effective in various continual learning scenarios. 
However, retraining old data and scaling for large-scale CIL can be challenging due to memory limitations. Additionally, privacy and safety issues may prevent data storage in certain situations \cite{parisi2019continual, li2020federated, appari2010information}.

\textbf{\textit{Parameter isolation-based methods}}, such as PackNet \cite{mallya2018packnet}, Piggyback \cite{mallya2018piggyback}, WA \cite{zhao2020maintaining}, SSIL \cite{ahn2021ss} and FOSTER \cite{wang2022foster}, allocate separate sets of parameters for different tasks. This strategy successfully reduces interference between tasks and allows for improved performance across multiple tasks. Despite these benefits, it comes with the trade-off of increased model complexity and potential inefficiencies in parameter usage.

Most of these methods stem from supervised learning, requiring old sample and model preservation, which can be cumbersome and raise privacy concerns \cite{de2021continual,zhu2021prototype}. Additionally, superior supervised learning techniques don't guarantee improvements of continual SSL models and may not be directly transferable.
\begin{figure*}[htb]
  \centering
  \includegraphics[width=0.9\linewidth]{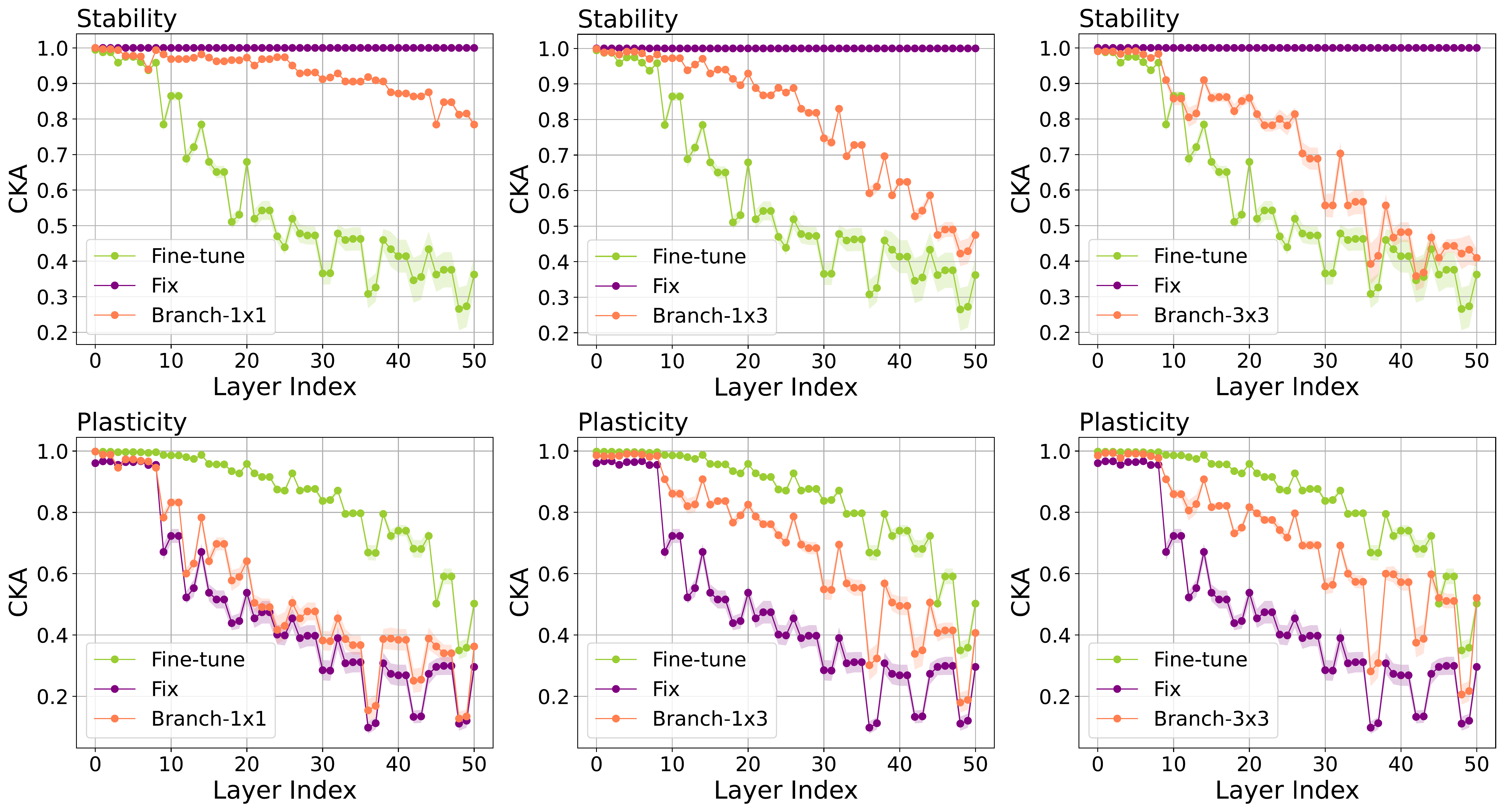}
  \vskip -0.1in
  \caption{Layer-wised stability and plasticity curves for Fixed, Fine-tuning, and Branch-tuning using 1x1, 1x3, and 3x3 structures. Branch-tuning achieves the best balance, as Fixed models exhibit low plasticity and Fine-tuning models have limited stability.}
  \vskip -0.1in
  \label{fig:2}
\end{figure*}
\subsection{Continual Self-Supervised Learning}
Continual self-supervised learning combines the benefits of self-supervised learning and continual learning, aiming to train SSL models using continuous data streams, with great potential for real-world scenarios.
Recently, a few continual SSL methods have been proposed. Co2L \cite{cha2021co2l} utilizes contrastive learning and self-supervised distillation, but it necessitates the use of labeled samples for distillation, thus not being entirely self-supervised. CaSSLe \cite{fini2022self} establishes benchmarks for continual SSL and presents a method that integrates self-supervised loss functions with unlabeled data distillation mechanisms using a predictor network.
Buffered SSL \cite{purushwalkam2022challenges} addresses inefficiency and catastrophic forgetting by introducing replay buffers and minimum redundancy (MinRed) buffers. Continual Barlow Twins \cite{marsocci2022continual} applies Continual SSL to remote sensing semantic segmentation, while Kaizen \cite{tang2023practical} reduces forgetting in feature extractors and classifiers using a customized loss function.

These approaches adopt techniques from prior supervised continual learning, including knowledge distillation and data replay, and show promising results. However, they also inherit limitations such as the requirement to store old data and models, and the use of complex and non-universal distillation or regularization protocols. In this work, we quantitatively analyze the stability and plasticity of SSL models in continual learning and propose a simple and direct method for continual SSL, named Branch-tuning.
\section{Preliminaries}
\label{sec: 3}
%
%
%
The continual SSL problem involves a sequence of $T$ tasks with disjoint datasets $\mathcal{D}=\{\mathcal{D}_1,\mathcal{D}_2,\dots,\mathcal{D}_T\}$. At phase $t$, the SSL model is updated using the training set in $\mathcal{D}_t$. We mainly consider the following two incremental settings:

\textbf{Class-incremental:} The dataset $\mathcal{D}_t$ contains images belonging to a set of classes $Y_t$, and $Y_t \cap Y_s = \emptyset$ for each other task $s \neq t$. The class labels are used only for splitting the dataset and are unknown to the model.

\textbf{Data-incremental:} In this case, $D_t \cap D_s = \emptyset$ for each other task $s \neq t$. No additional constraints are imposed on the classes. In practice, the whole dataset is shuffled and then partitioned into $T$ tasks. Each task can potentially contain all the classes.

For analysis purposes, we represent the SSL model with two parts: a feature extractor $\mathcal{F}$ and a projector $g$. The feature extractor, parameterized by $\theta$, encodes the input into a representation as $z=\mathcal{F}(x;\theta)$. The representation is then passed through the projector to obtain $g(z)$ for $x$, which is used to calculate the similarity between different representations. At time $t$, the general goal is to update the model $\Theta_t$ from the old $\Theta_{t-1}$, and afterward, the projector g is discarded.

\section{Investigate Stability and Plasticity in Continual SSL}
\label{sec:4}
\subsection{Centered Kernel Alignment}
\label{sec:4.1}
Assessing stability and plasticity in continual SSL requires a quantifiable understanding of neural network behavior in these conditions. To this end, we introduce the Centered Kernel Alignment (CKA) metric \cite{cortes2012algorithms,kornblith2019similarity}, an efficient method for comparing feature representation similarity between two networks.

Given two sets of feature representations $X \in \mathbb{R}^{n \times d_x}$ and $Y \in \mathbb{R}^{n \times d_y}$, where $n$ is the number of samples and $d_x$ and $d_y$ are the feature dimensions, we first center the feature representations by subtraction of their means:
\begin{equation}
\begin{aligned}
\tilde{X} &= X - \frac{1}{n} X \mathbf{1}_{n \times n}, \\
\tilde{Y} &= Y - \frac{1}{n} Y \mathbf{1}_{n \times n},
\end{aligned}
\end{equation}
where $\mathbf{1}_{n \times n}$ is an $n \times n$ matrix with all elements equal to $\frac{1}{n}$. Following this, we compute the centered Gram matrices, $K_X$ and $K_Y$, as the inner products of the centered representations:
\begin{equation}
	\begin{aligned}
	K_X &= \tilde{X} \tilde{X}^T, \\
	K_Y &= \tilde{Y} \tilde{Y}^T.
	\end{aligned}
\end{equation}
Finally, we calculate the CKA metric by evaluating the normalized Frobenius inner product of the centered Gram matrices:
\begin{equation}
\text{CKA}(X, Y) = \frac{\langle K_X, K_Y \rangle_F}{|K_X|_F |K_Y|_F},
\end{equation}
where $\langle \cdot, \cdot \rangle_F$ denotes the Frobenius inner product, and $|\cdot|_F$ represents the Frobenius norm. The CKA metric varies between 0 and 1, with 0 signifying no similarity and 1 indicating identical feature representations.

\subsection{Quantify Stability and Plasticity}
\label{sec:4.2}

To evaluate stability and plasticity during the continual SSL process, we measure the similarity of new and old data features between the continuously learned model and the joint-trained model.
By merging and retraining all previous and current data at each incremental task, the joint-trained model achieves optimal stability and plasticity. The definitions of stability ($S$) and plasticity ($P$) at stage $i$ in the continual SSL process are:

\begin{equation}
\begin{aligned}
%
%
S_i&= \mathit{CKA}({\mathcal{F}_{i-1}}(X_{0}^{i-1}), \mathcal{F}_{i}(X_{0}^{i-1})),\\
P_i&= \mathit{CKA}(\widehat{\mathcal{F}_{i}}(X_{i-1}^{i}), \mathcal{F}_{i}(X_{i-1}^{i})).
\end{aligned}
\end{equation}
Here, $\widehat{\mathcal{F}_{i}}$ and $\mathcal{F}_{i}$ refer to the jointly trained model and the continually learned model at stage $i$, respectively. $X_{i}^{j}$ represents the training data from stage $i$ to $j$ in the continual learning process.

We conducted a two-stage continual SSL experiment on the CIFAR-100 \cite{krizhevsky2009learning} dataset using a ResNet-18 \cite{he2016deep} backbone and the SimCLR model, and performed an in-depth layer-wise analysis of stability and plasticity. Given a pair of incremental feature extractors $\mathcal{F}_0$ and $\mathcal{F}_1$ trained in a continual SSL model, we calculated stability $S_1$ and plasticity $P_1$ for all intermediate variable layers using the $\mathcal{D}_0$ and $\mathcal{D}_1$ validation sets. We extracted activations from all convolution, batch normalization, and residual block layers of ResNet-18, resulting in two sets of stability and plasticity, each with 51 elements \cite{kim2023stability}.


Figure \ref{fig:2} presents layer-wise stability and plasticity between $\mathcal{F}_0$ and $\mathcal{F}_1$, visualizing three methods: 1) the fixed model, 2) the fine-tuned model, and 3) our method (not discussed for now). Layer-wise stability corresponds to the similarity between a model's layer and its previous state on old data, with higher stability maximizing the preservation of the model's capabilities. On the other hand, plasticity measures the similarity between a model's layer features on new data and an offline-trained joint model, where higher plasticity suggests that the model's learning ability on new data is approaching the optimal effect.

Examining the fixed model, we observed absolute stability but significantly lower plasticity, indicating a considerable difference in adaptability to new data compared to retraining. For the fine-tuned model, stability declines noticeably as parameters important for old tasks are updated during adaptation to new tasks, implying changes in feature distribution on old data. We observed a trend of relatively high CKA in early layers that deteriorates in latter layers, which aligns with the notion that early layers learn low-level and widespread features, while higher layers learn more class-specific information \cite{lecun2015deep,goodfellow2016deep,yosinski2014transferable}. The fine-tuned model's layer similarity to the retrained model on new data is higher than the fixed model, showcasing better adaptability to new data.

This phenomenon confirms the continual SSL dilemma: Fine-tuning a model often leads to insufficient stability and forgetting, while enforcing stability limits the model's adaptability to new data. The key challenge is striking a balance between stability and plasticity, allowing the model to effectively learn from new data while retaining knowledge from previous tasks.
\begin{table}[t]

   \caption{Linear-Probe accuracy for old and new classes when fixing or Fine-tuning BN and convolutional layers in models.}
    \label{table:1}
       \vskip -0.1in
 \small \centering 
 \renewcommand\tabcolsep{6.5pt}
 \renewcommand{\arraystretch}{1.3}
\begin{tabular}{cc|ccc}
\toprule[1.3pt]
\multicolumn{2}{c|}{\textbf{Strategy}} & \multicolumn{3}{c}{\textbf{Linear-Probe Acc(\%)}} \\ \hline
Conv Layers          & BN Layers          & Old Task       & New Task       & Average        \\ \hline
Fix                 & Fix                & \textbf{51.7} & 36.5          & 44.1          \\
Fix                 & Fine-tune          & 49.5          & 38.1          & 43.8           \\
Fine-tune           & Fine-tune          & 47.4          & \textbf{42.9} & 45.2         \\
\rowcolor{mygray}Fine-tune           & Fix                & 49.7          & 41.4          & \textbf{45.6} \\ 
  \bottomrule[1.3pt]
\end{tabular}
\end{table}

\begin{figure}[t]
  \centering
  \includegraphics[width=0.8\linewidth]{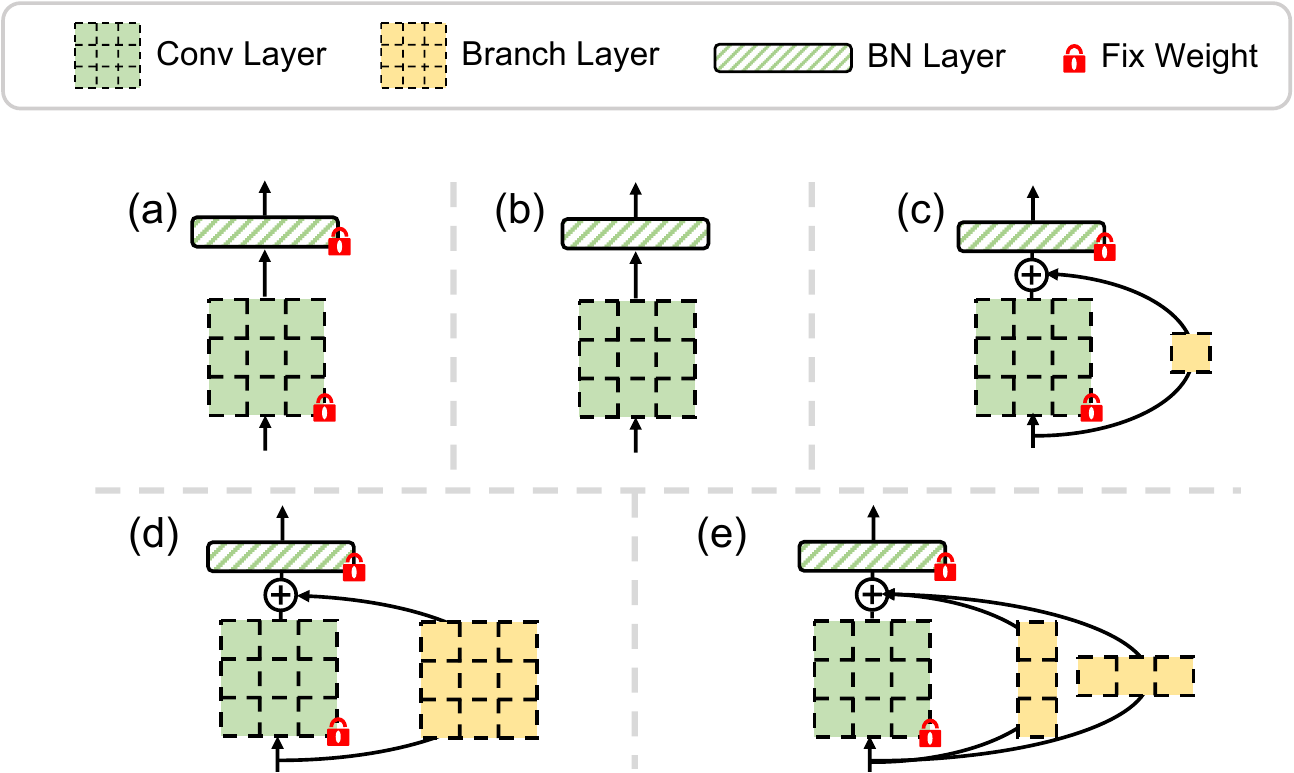}
  \caption{Difference between (a) a fixed model, (b) Fine-tuning, (c), (d), and (e) our method with 1x1, 1x3, and 3x3 branches.}
  \vskip -0.1in
  \label{fig:3}
   \vskip -0.1in
\end{figure}
\begin{figure*}[t]
  \centering
  \includegraphics[width=1.0\linewidth]{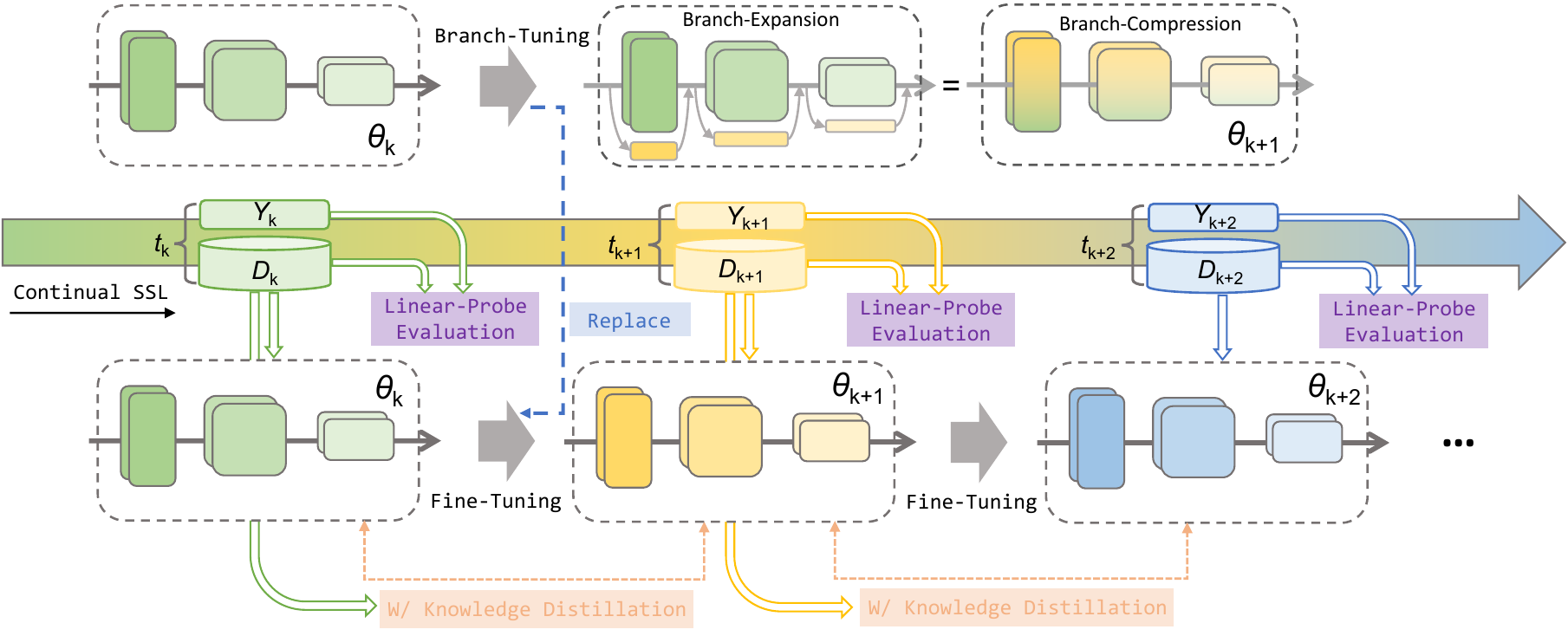}
  \caption{Overview of Branch-tuning, a method to train SSL models in continuous streams of unlabeled data, achieving a balance between stability and plasticity. Our method comprises Branch-Expansion and Branch-Compression and can be applied to various SSL models without modifications. Model performance is assessed using Linear-Probe evaluation with labeled data.}
  \vskip -0.1in
  \label{fig:4}
\end{figure*}
\subsection{Understand Stability and Plasticity in CNNs}
In this study, we aim to identify the key components within Convolutional Neural Networks (CNNs) that impact stability and plasticity in continual SSL. There are two extreme strategies for model adaptation in continual learning: (1) maintaining a fixed model on new tasks, and (2) Fine-tuning the model on new tasks without constraints. To understand the influence of different components, we focus on the role of trainable modules within CNNs, specifically convolutional layers (Conv layers) and batch normalization layers (BN layers), and how they affect stability and plasticity in the learning process.

Following the two-stage continual SSL experiments in Section \ref{sec:4.2}, we examined the linear-probe accuracy of Conv and BN layers under these two extreme scenarios on both old and new classes. This analysis provides an intuitive reflection of their contributions to stability and plasticity, corresponding to the performance on old and new data, respectively.
Surprisingly, as demonstrated in Table \ref{table:1}, fixing BN layers significantly improves model stability, whereas Conv layers play a more prominent role in model plasticity. Based on these findings, we recommend fixing BN layers and adjusting Conv layers during the continual SSL process.

However, direct Fine-tuning of Conv layers negatively impacts stability, leading to the forgetting of prior knowledge. To address this issue, as illustrated in Figure \ref{fig:3}, we propose an approach that finds a middle ground between fixing and Fine-tuning in order to balance stability and plasticity. We tested three different branch structures, denoted as Branch 1x1, Branch 1x3, and Branch 3x3.  As demonstrated in Figure \ref{fig:2}, following the two-stage continual SSL experiments described in Section \ref{sec:4.2}, Our method outperforms Fine-tuning in terms of stability and surpasses a fixed model in terms of plasticity, leading to a more balanced result. These insights provide the primary inspiration for our method.

\section{Method: Branch-tuning}
\label{sec:5}
\subsection{Overview of the Framework}
In this section, we introduce our method, Branch-tuning, which consists of two components: Branch Expansion and Branch Compression. It is compatible with most existing SSL and continual SSL methods, e.g., BYOL \cite{grill2020bootstrap}, MoCoV2 \cite{he2020momentum}, SimCLR \cite{chen2020simple},  BarlowTwins \cite{zbontar2021barlow}, CaSSLe \cite{fini2022self}, PFR \cite{gomez2022continually}, and POCON \cite{gomez2024plasticity}. Branch Expansion is designed to balance stability and plasticity when learning new knowledge. 
Specifically, during training on a batch of new data, we fix the existing BN and Conv layers in the model to mitigate the forgetting of prior knowledge. We then introduce a new branch layer alongside the Conv layer, and train the parameters of the branch layer using the new data. This method is referred to as Branch Expansion. After that, Branch Compression equivalently converts the branch into the original network through reparameterization, keeping the network structure unchanged during the continual SSL process. 

Figure \ref{fig:4} illustrates our method. The innovations mainly lie in two aspects. First, Branch-tuning can be applied to existing SSL models without modification, by replacing the standard CNN network structure. It eliminates the need to design different knowledge distillation or regularization rules for various SSL models, which can be complex and time-consuming and potentially lack generalizability.
Second, our method does not rely on retaining old data or models, which may pose privacy concerns and significantly increase storage overhead. Branch-tuning does not require pre-set hyperparameters, offering excellent generalizability and simplicity.

\subsection{Branch Expansion}
Self-supervised learning trains a feature extractor $\mathcal{F}$ using two data augmentations of images $(x^A, x^B)$, by pulling the representations of positive pairs closer in the feature space. The process can be described by the following algorithm:
\begin{algorithm}[t]
\caption{Process of Self-supervised Learning}
\begin{algorithmic}[1]
\STATE \textbf{Input:} Dataset $\mathcal{D}$, feature extractor $\mathcal{F}$, project head $g$
\FOR{batch data augmentation$(x^A, x^B)$ in $\mathcal{D}$}
\STATE Obtain representations: $h^A \gets \mathcal{F}(x^A)$, $h^B \gets \mathcal{F}(x^B)$
\STATE Obtain projection: $z^A \gets g(h^A)$, $r^B \gets g(h^B)$
\STATE Compute loss: $\mathcal{L}_{SSL} \gets \mathcal{L}(z^A, z^B)$
\STATE Update $\mathcal{F}$ and $g$ to minimizing $\mathcal{L}_{SSL}$
\ENDFOR
\STATE \textbf{Output:} Feature extractor $\mathcal{F}(\cdot)$, and throw away $g(\cdot)$
\end{algorithmic}
\end{algorithm}
For example, the calculations of self-supervised learning loss based on InfoNCE and Mean Squared Error (MSE) are as follows:
\begin{equation}
\mathit{\mathcal{\ell}_{InfoNCE}}(\rm{z}^A_i, \rm{z}^B_j) = - \log \frac{\exp(\rm{sim}( z^A_i, z^B_j) / \tau)}{\sum_{k\neq i}\exp(\rm{sim}( z^A_i, z^B_k) / \tau)},
\end{equation}
\begin{equation}
\mathit{\mathcal{L}_{InfoNCE}}= \frac{1}{2N} \sum_{k=1}^{N} \left[\ell(\rm{z}^A_{2k-1}, \rm{z}^B_{2k}) + \ell(\rm{z}^B_{2k},\rm{z}^A_{2k-1}) \right],
\end{equation}
\begin{equation}
\mathit{\mathcal{L}_{MSE}} = |z^A - z^B|^2,
\end{equation}
where $\rm{sim}$ is a similarity function, $N$ denotes batch size, and $\tau$ controls distribution sharpness in InfoNCE loss.

The primary goal of the Branch Expansion is to maintain a balance between stability and plasticity by training a new branch convolutional layer with new data while keeping the original convolutional layer unchanged. This approach results in better stability and plasticity trade-offs compared to directly fixing or Fine-tuning the convolutional layer.
Branch Expansion can be easily applied to self-supervised models by replacing the feature extractor $\mathcal{F}$. Specifically, for the feature extractor at stage $i$, $\mathcal{F}_i$, which contains several convolutional layers denoted as $[f_i^1, f_i^2, \dots, f_i^M]$, we replace these convolutional layers $f_i^j$ to obtain $[\bar{f_i^1}, \bar{f_i^2}, \dots, \bar{f_i^M}]$ and get the Branch Expansion feature extractor $\bar{\mathcal{F}}_i$ as follows:
\begin{equation}
\bar{f_i^j}(x_i) = \text{StopGrad}(f_i^j(x_i)) + h_i^j(x_i).
\end{equation}
In this context, $h_i^j$ refers to the $j$-th convolution layer branch in stage $i$, which is utilized to acquire new knowledge. The $\rm{StopGrad}$ function stops the gradient flow, preserving the old knowledge by fixing the previous convolution layer.
The self-supervised loss function for Branch Expansion on data $\mathcal{D}_i$ at stage $i$ can be defined as:
\begin{equation}
\mathit{\mathcal{L}_{SSL}} = \mathcal{L}(g(\bar{F}_i(X^{A}_i)), g(\bar{F}_i(X^{B}_i))).
\end{equation}

Our method eliminates the need to preserve old models and data, reducing storage overhead as the number of incremental tasks grows. Furthermore, it avoids the complex process of designing distillation loss for various SSL models, increasing its adaptability across different models.

\subsection{Branch Compression}
\begin{figure}[t]
  \centering
  \vskip -0.1in
  \includegraphics[width=0.8\linewidth]{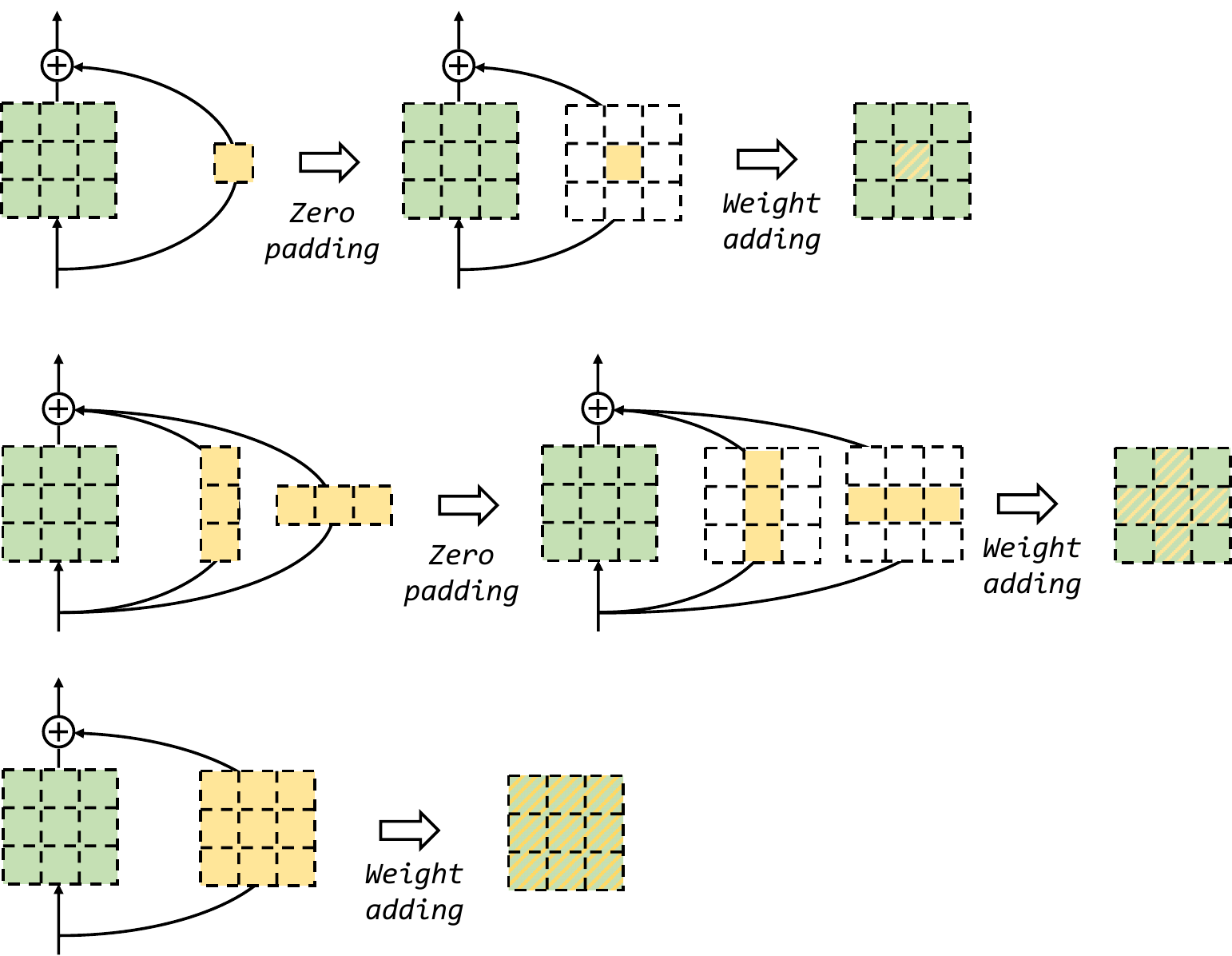}
  \vskip -0.1in
  \caption{Illustration of Branch Compression with three branch structures, 1x1, 1x3, and 3x3.}
  \vskip -0.1in
  \label{fig:5}
\end{figure}
In the continual SSL process, introducing new branches while learning new data can result in a bloated network structure and continuously increasing parameters. To address this issue, we employ Branch Compression using reparameterization, which transforms two convolutional layers of different sizes into a single layer, as demonstrated in Figure \ref{fig:5}.

To be specific, reparameterization relies on the fact that the convolution operation is a linear operation, which satisfies the commutative property for addition. First, we need to align the kernel sizes of the branch layer and the convolution layer. This is achieved by padding the smaller convolution kernels with zeros to make them equivalent to the larger kernels. For instance, in Figure \ref{fig:5}, a 1x1 or 1x3 convolution $h_{i}^{j}$ can be padded to match a 3x3 convolution $\hat{h}_{i}^{j}$.
After that, by utilizing the commutative property of addition in the convolution operation, the sum of the outputs of two convolutions can be considered equivalent to the output of a new convolution layer with combined kernel values. This process is shown below:
\begin{equation}
	\begin{aligned}
		\text{conv}(x, w_{ij}) + \text{conv}(x, w'_{ij})&=\text{conv}(x, w_{ij} + w'_{ij}) \\
														&= \text{conv}(x, \hat{w}_{ij}).
	\end{aligned}
\end{equation}
Where $w_{ij}$ are the weights of $f_{i}^{j}$, $w'_{ij}$ are the weights of $h_{i}^{j}$, and $\hat{w}_{ij}$ are the weights of $\hat{f}_{ij}$, which are obtained after reparameterization of the $j$-th convolution layer upon completing the learning of the $i$-th incremental task. $\text{conv}$ represents the convolution operation. By performing the above process for each convolution layer in the network, we can obtain a feature extractor $\hat{\mathcal{F}}$ with a consistent structure as the original network, which can be used for continuous learning in the next stage.
%
%

%
%

Finally, we present the Branch-tuning method for continual SSL, as shown in Algorithm \ref{alg:2}, which consists of Branch Expansion and Branch Compression.

Branch-tuning can be directly applied to different SSL models without modification, eliminating the need to store old models and data, and requiring no hyperparameters to be set. By balancing stability and plasticity in the continual SSL process, it enables models to adaptively learn new knowledge in the data stream of real-world scenarios.

\begin{algorithm}[H]
\renewcommand{\algorithmicrequire}{\textbf{Input:}} 
\renewcommand{\algorithmicensure}{\textbf{Output:}} 
\caption{Process of Branch-tuning}
\label{alg:2}
\begin{algorithmic}[1]
\STATE \textbf{Input:} Dataset $\mathcal{D}$, feature extractor $\mathcal{F}$, project head $g$
\FOR{each incremental task $i$}
\STATE \textbf{$\triangleright$ Apply Branch Expansion}
\FOR{batch data augmentation $(x^A, x^B)$ in $\mathcal{D}_i$}
\STATE $h^A \gets \bar{\mathcal{F}_i}(x^A)$, $h^B \gets \bar{\mathcal{F}_i}(x^B)$
\STATE $z^A \gets g(h^A)$, $r^B \gets g(h^B)$
\STATE $\mathcal{L}_{SSL} \gets \mathcal{L}(z^A, z^B)$

\STATE Update $\bar{\mathcal{F}_i}$ and $g$ to minimize $\mathcal{L}_{SSL}$
\ENDFOR
\STATE \textbf{$\triangleright$ Apply Branch Compression}
\STATE get feature extractor $\mathcal{F}_i$ from $\bar{\mathcal{F}_i}$ using re-parametrized convolutions
\ENDFOR
\STATE \textbf{Output:} Feature extractor $\mathcal{F}(\cdot)$, and discard $g(\cdot)$
\end{algorithmic}
\end{algorithm}

\section{Experiments}
\label{sec:6}
\subsection{Evaluation Metrics}
The experimental metrics for evaluating continual SSL models are based on the state-of-the-art method CaSSLe \cite{fini2022self}. We use three main metrics to assess our approach:

\noindent\textbf{Linear-Probe Evaluation:} In the context of continual SSL, Linear-Probe accuracy is determined by training a linear classifier on a dataset's training set and evaluating its accuracy on a specific task. After completing incremental learning at stage $t$, we test the Linear-Probe accuracy $A_{i,t}$ for each task dataset $\mathcal{D}_i$. The average accuracy across all tasks following the final training task is calculated as $A = \frac{1}{N} \sum{i=1}^{N} A_{i,T}$.


\noindent\textbf{Forgetting:} This metric quantifies the amount of information the model has forgotten about previous tasks. It is formally defined as:
$F = \frac{1}{T-1} \sum_{i=1}^{T-1} \max_{t \in {1, \dots, T}} (A_{t,i} - A_{T,i})$.

\noindent\textbf{Forward Transfer:} This metric evaluates how much the learned representations are helpful in learning new tasks. The transfer measure is given by $\text{FT} = \frac{1}{T-1} \sum_{i=2}^{T} (A_{i-1,i} - R_i)$, where $R_i$ is the linear evaluation accuracy of a random network on task $i$.

\subsection{Experimental Setup}

\noindent\textbf{Datasets.} We conduct experiments on four popular datasets. Continuous SSL experiments are performed on \textit{ImageNet-100} \cite{deng2009imagenet},\textit{CIFAR-100} \cite{krizhevsky2009learning}, and \textit{TinyImageNet}, which include class incremental and data incremental settings. By default, these datasets are randomly shuffled and evenly divided into 5 incremental tasks. Transfer learning experiments are conducted on the \textit{RAF-DB}  \cite{li2017reliable} dataset, a real-world facial expression recognition dataset, to demonstrate the effectiveness of our method in real-world scenarios.

\begin{figure*}[t]
  \centering
  \includegraphics[width=1.0\linewidth]{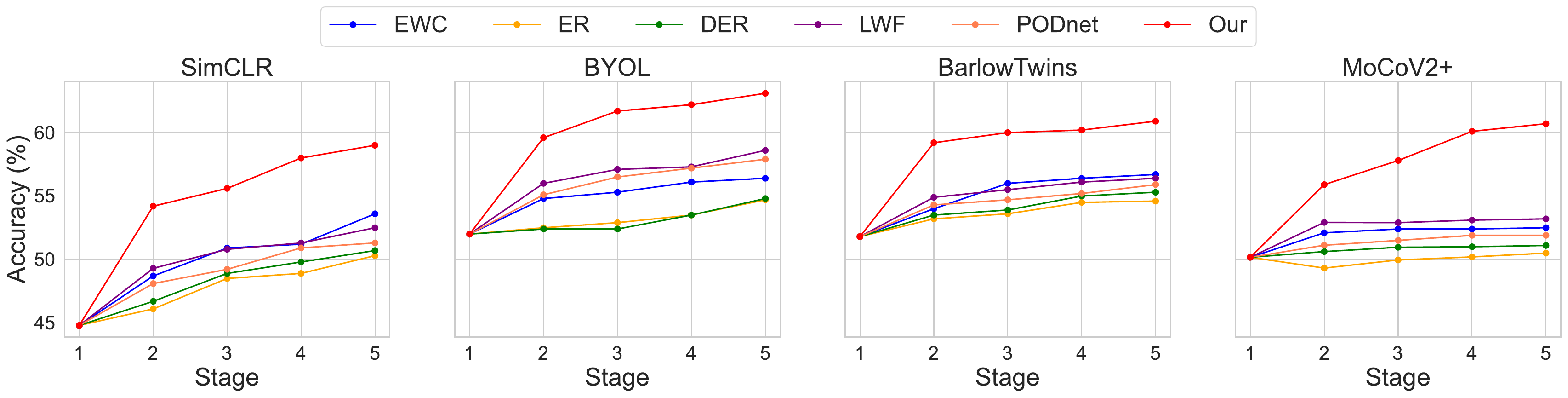}
  \vskip -0.1in
  \caption{Linear-Probe evaluation on CIFAR-100 in 5 tasks class-incremental setting. Performance at each stage is compared between the continual SSL (CaSSLe + BT) and recent supervised CIL methods \cite{robins1995catastrophic,kirkpatrick2017overcoming,li2017learning,douillard2020podnet,buzzega2020dark}. Continual SSL methods significantly outperform Supervised CIL methods.}
  \vskip -0.1in
  \label{fig:6}
\end{figure*}

\begin{table*}[t]

      \caption{Linear-Probe accuracy on CIFAR-100 after 5-task class-incremental SSL, compared with CIL methods. Baselines results are come from \cite{fini2022self}.}
    \label{table:2}
    \vskip -0.1in
 \small \centering 
 \renewcommand\tabcolsep{3pt}
 \renewcommand{\arraystretch}{1.3}
\begin{tabular}{lcccccccccccc}
\toprule[1.3pt] 
\multirow{2}{*}{\textbf{Method}} & \multicolumn{3}{c}{\textbf{SimCLR}}          & \multicolumn{3}{c}{\textbf{BYOL}}            & \multicolumn{3}{c}{\textbf{Barlow Twins}}    & \multicolumn{3}{c}{\textbf{MoCoV2+}}         \\
                                & A ($\uparrow$)                              & F ($\downarrow$)                         & T  ($\uparrow$)                            & A ($\uparrow$)                            & F ($\downarrow$)                         & T   ($\uparrow$)              & A ($\uparrow$)                              & F ($\downarrow$)                         & T  ($\uparrow$) & A ($\uparrow$)                              & F ($\downarrow$)                         & T  ($\uparrow$)            \\ \hline
\textit{\textbf{Supervised CIL}} & \multicolumn{12}{c}{}                                                                                                                                                                     \\
EWC \cite{kirkpatrick2017overcoming}                             & 53.6          & 0.0          & 33.3          & 56.4          & 0.0          & 39.9          & 56.7          & 0.2          & 39.1          & 52.5          & 0.0          & 35.0          \\
ER \cite{robins1995catastrophic}                               & 50.3          & 0.1          & 32.7          & 54.7          & 0.4          & 36.3          & 54.6          & 3.0          & 39.4          & 50.5          & 0.1          & 33.7          \\
DER \cite{buzzega2020dark}                              & 50.7          & 0.4          & 33.2          & 54.8          & 1.1          & 36.7          & 55.3          & 2.5          & 39.6          & 51.1          & 0.0          & 34.3          \\
LWF \cite{li2017learning}                     & 52.5          & 0.2          & 33.8          & 58.6          & 0.2          & 41.1          & 56.4          & \textbf{0.2} & 40.1          & 53.2          & 0.0          & 36.7          \\
PODnet \cite{douillard2020podnet}                              & 51.3          & \textbf{0.1} & 33.8          & 57.9          & 0.0          & 41.1          & 55.9          & 0.3          & 40.3          & 51.9          & 0.1          & 35.3          \\ \hline
\textit{\textbf{Continual SSL}}  & \multicolumn{12}{c}{}                                                                                                                                                                     \\
Branch-tuning                    & 53.2          & 0.4          & 34.2          & 58.0          & 0.0          & 38.4          & 57.2          & 0.4          & 40.3          & 52.7          & 0.1          & 35.8          \\
CaSSLe +BT(our)                 & \textbf{59.0} & 0.2          & \textbf{36.9} & \textbf{63.1} & \textbf{0.0} & \textbf{43.0} & \textbf{60.9} & 0.4          & \textbf{43.0} & \textbf{60.7} & \textbf{0.0} & \textbf{40.1} \\
Offline                          & 65.8          &              &               & 70.5          &              &               & 70.9          &              &               & 69.9          &              &               \\ 
\bottomrule[1.3pt]
\end{tabular}
\vskip -0.1in
\end{table*}

\vspace*{5pt}
\noindent\textbf{Comparison Methods.} Most supervised CIL methods rely on labeled data, which is not applicable to continual SSL. However, some methods can be adapted to the continual SSL setting with minor adjustments \cite{fini2022self}. We have chosen to compare three types of baseline methods and SOTA methods: regularization-based methods such as EWC \cite{kirkpatrick2017overcoming} and "Less-Forget" \cite{li2017learning}, and exemplar-based methods (e.g., PODnet \cite{douillard2020podnet}, ER \cite{robins1995catastrophic}, DER \cite{buzzega2020dark}).
We also compared with recent state-of-the-art continual SSL methods, including CaSSLe \cite{fini2022self}, PFR \cite{gomez2022continually}, and POCON \cite{gomez2024plasticity}. We applied branch-tuning (BT) to these methods, denoted as, for example, CaSSLe +BT.
Lastly, we did not consider methods based on VAEs \cite{achille2018life,rao2019continual} due to their poor performance when dealing with large-scale data.

\vspace*{5pt}
\noindent\textbf{Implementation Details.} Our implementation is based on two well-established libraries: Solo-learn \cite{da2022solo}, a widely-used SSL library, and CaSSLe \cite{fini2022self}, a SOTA continual SSL model. The number of epochs per task is set as follows: 500 for CIFAR-100, 400 for ImageNet-100, and 500 for TinyImageNet. We employ a ResNet-18 \cite{he2016deep} backbone with a batch size of 256 and use the LARS \cite{you2017large} optimizer. To ensure a fair comparison, the offline version of each method, which serves as an upper bound, is trained for the same number of epochs as its continual counterpart.
In the transfer experiments on RAF-DB, we employ encoders trained on CIFAR-100 in a class incremental setting and evaluate their performance using Linear-Probe. The transfer training is conducted for 50 epochs with the SGD \cite{krizhevsky2012imagenet} optimizer.

\subsection{Experimental Results}
In this section, we discuss the results of experiments on class-incremental, data-incremental, and transfer learning settings. 

\vspace*{5pt}
\noindent\textbf{Comparison with CIL methods.}
Continual SSL faces several challenges when leveraging Class Incremental Learning (CIL) methods that are based on supervised learning. These issues include \textbf{(1)} reliance on labeled samples, which hinders application to SSL models; \textbf{(2)} the need to save old models and samples for knowledge distillation, and \textbf{(3)} potential conflicts of certain regularization losses with SSL models. As demonstrated in Table \ref{table:2} and Figure \ref{fig:6}, Exemplar-based methods such as ER \cite{robins1995catastrophic} and DER \cite{buzzega2020dark}, which have been effective in supervised CIL, are not useful in continual SSL as they rely on saving labeled samples to prevent forgetting. This is not possible when sample annotations are unavailable. Additionally, SSL requires more training epochs than supervised learning, which can lead to overfitting on a small number of old samples. ER and DER provide minimal improvement compared to direct Fine-tuning, whereas Branch-tuning can boost accuracy significantly without requiring saving old samples/models or complex knowledge distillation.
Some Knowledge Distillation (KD)-based supervised CIL methods, such as EWC \cite{kirkpatrick2017overcoming}, LWF \cite{li2017learning}, and PODnet \cite{douillard2020podnet}, show performance improvements under continual SSL. However, their distillation losses may conflict with SSL models, causing information loss when paired with Barlow Twins and reducing accuracy. On the contrary, Branch-tuning, combined with the SOTA distillation method in continual SSL, CaSSLe, can boost accuracy and is compatible with most SSL models, achieving a significant performance increase( 4.2-7.5\%) over supervised CIL methods.

\begin{table}[htb]

   \caption{Linear-Probe evaluation on CIFAR-100 and ImageNet-100 after 5-task class-incremental SSL. Metrics assessed consist of average accuracy (A), forgetting rate (F), and Forward Transfer (T).}
    \label{table:3}
       \vskip -0.1in
 \small \centering 
 \renewcommand\tabcolsep{1.5pt}
 \renewcommand{\arraystretch}{1.3}
\begin{tabular}{clcccccc}
\toprule[1.3pt] 
                                                                         &                                              & \multicolumn{3}{c}{\textbf{CIFAR-100}}                                                                                & \multicolumn{3}{c}{\textbf{ImageNet-100}}                                                                            \\ \cline{3-8} 
\multirow{-2}{*}{\textbf{Method}}                                        & \multirow{-2}{*}{\textbf{Strategy}}          & A ($\uparrow$)                              & F ($\downarrow$)                         & T  ($\uparrow$)                            & A ($\uparrow$)                            & F ($\downarrow$)                         & T   ($\uparrow$)                           \\ \hline
                                                                         & Fine-tuning                                  & 48.9                                   & 1.0                                  & 33.5                                  & 61.5                                  & 8.1                                  & 40.3                                  \\
                                                                         & \cellcolor{mygray}Branch-tuning        & \cellcolor{mygray}\textbf{53.2} & \cellcolor{mygray}\textbf{0.4} & \cellcolor{mygray}\textbf{34.2} & \cellcolor{mygray}\textbf{62.8} & \cellcolor{mygray}\textbf{5.8} & \cellcolor{mygray}\textbf{42.2} \\  
                                                                         & CaSSLe \cite{fini2022self}                          & 58.3                                   & 0.2                                  & 36.4                                  & 68.0                                  & 2.2                                  & 45.8                                  \\
                                                                         & \cellcolor{mygray}CaSSLe +BT & \cellcolor{mygray}\textbf{59.0}  & \cellcolor{mygray}\textbf{0.2} & \cellcolor{mygray}\textbf{36.9} & \cellcolor{mygray}\textbf{68.5} & \cellcolor{mygray}\textbf{1.9} & \cellcolor{mygray}\textbf{46.3} \\  
\multirow{-5}{*}{SimCLR}                                                 & Offline                                      & 65.8                                   & -                                    & -                                     & 77.5                                  & -                                    & -                                     \\ \hline
                                                                         & Fine-tuning                                  & 52.7                                   & 0.1                                  & 35.9                                  & 66.0                                  & 2.9                                  & 43.2                                  \\
                                                                         & \cellcolor{mygray}Branch-tuning        & \cellcolor{mygray}\textbf{58.0} & \cellcolor{mygray}\textbf{0.0} & \cellcolor{mygray}\textbf{38.4} & \cellcolor{mygray}\textbf{66.8} & \cellcolor{mygray}\textbf{1.0} & \cellcolor{mygray}\textbf{44.5} \\  
                                                                         & CaSSLe \cite{fini2022self}                         & 62.2                                   & 0.0                                  & 42.2                                  & 66.4                                  & 1.1                                  & 46.6                                  \\
                                                                         & \cellcolor{mygray}CaSSLe +BT & \cellcolor{mygray}\textbf{63.1}  & \cellcolor{mygray}0.0          & \cellcolor{mygray}\textbf{43.5}   & \cellcolor{mygray}\textbf{67.5} & \cellcolor{mygray}\textbf{0.8} & \cellcolor{mygray}\textbf{48.3} \\  
\multirow{-5}{*}{BYOL}                                                   & Offline                                      & 70.5                                   & -                                    & -                                     & 80.3                                  & -                                    & -                                     \\ \hline
                                                                         & Fine-tuning                                  & 54.3                                   & 0.4                                  & 39.2                                  & 63.1                                  & 10.7                                 & 44.4                                  \\
                                                                         & \cellcolor{mygray}Branch-tuning        & \cellcolor{mygray}\textbf{57.2} & \cellcolor{mygray}0.4          & \cellcolor{mygray}\textbf{40.3} & \cellcolor{mygray}\textbf{66.4} & \cellcolor{mygray}\textbf{5.3} & \cellcolor{mygray}\textbf{45.9} \\  
                                                                         & CaSSLe \cite{fini2022self}                         & 60.4                                   & 0.4                                  & 42.2                                  & 68.2                                  & 1.3                                  & 47.9                                  \\
                                                                         & \cellcolor{mygray}CaSSLe +BT & \cellcolor{mygray}\textbf{60.9}  & \cellcolor{mygray}0.4          & \cellcolor{mygray}\textbf{43.0}   & \cellcolor{mygray}\textbf{69.1} & \cellcolor{mygray}\textbf{1.0} & \cellcolor{mygray}\textbf{48.4} \\  
\multirow{-5}{*}{\begin{tabular}[c]{@{}c@{}}Barlow\\ Twins\end{tabular}} & Offline                                      & 70.9                                   & -                                    & -                                     & 80.4                                  & -                                    & -                                     \\ \hline
                                                                         & Fine-tuning                                  & 47.3                                   & 0.2                                  & 33.4                                  & 62.0                                    & 8.4                                  & 41.6                                  \\
                                                                         & \cellcolor{mygray}Branch-tuning        & \cellcolor{mygray}\textbf{52.7} & \cellcolor{mygray}\textbf{0.1} & \cellcolor{mygray}\textbf{35.8} & \cellcolor{mygray}\textbf{65.9} & \cellcolor{mygray}\textbf{4.3} & \cellcolor{mygray}\textbf{44.2} \\  
                                                                         & CaSSLe \cite{fini2022self}                         & 59.5                                   & 0.0                                  & 39.6                                  & 68.8                                  & 1.5                                  & 46.8                                  \\
                                                                         & \cellcolor{mygray}CaSSLe +BT & \cellcolor{mygray}\textbf{60.7}  & \cellcolor{mygray}0.0          & \cellcolor{mygray}\textbf{40.1} & \cellcolor{mygray}\textbf{69.4} & \cellcolor{mygray}\textbf{1.3} & \cellcolor{mygray}\textbf{48.1} \\  
\multirow{-5}{*}{MoCoV2+}                                                & Offline                                      & 69.9                                   & -                                    & -                                     & 77.5                                  & -                                    & -                                     \\ 
\bottomrule[1.3pt]
\end{tabular}
\end{table}

\begin{table}[htb]

   \caption{Linear-Probe accuracy on CIFAR-100, TinyImageNet, and ImageNet-100 after different task class-incremental SSL. Datasets are divided across 4 to 100 tasks, experimental setup, and baseline results come from POCON \cite{gomez2024plasticity}.}
    \label{table:pocon}
    \vskip -0.1in
 \small \centering 
 \renewcommand\tabcolsep{5pt}
 \renewcommand{\arraystretch}{1.3}
\begin{tabular}{clccccc}
\toprule[1.3pt] 
                                   &                                       & \multicolumn{5}{c}{\textbf{Tasks}}                                                                                                                                                                    \\ \cline{3-7} 
\multirow{-2}{*}{\textbf{Dataset}} & \multirow{-2}{*}{\textbf{Strategy}}   & 4                                     & 10                                    & 20                                    & 50                                    & 100                                   \\ \hline
                                   & Fine-tuning                           & 54.8                                  & 50.9                                  & 45.0                                  & 38.0                                  & 27.0                                  \\
                                   & \cellcolor{mygray}Branch-tuning & \cellcolor{mygray}57.8 & \cellcolor{mygray}52.2          & \cellcolor{mygray}47.3          & \cellcolor{mygray}42.5          & \cellcolor{mygray}39.5          \\
                                   & PFR \cite{gomez2022continually}                                  & 59.7                                  & 54.3                                  & 44.8                                  & 46.5                                  & 43.3                                  \\
                                   & \cellcolor{mygray}PFR +BT        & \cellcolor{mygray}59.5 & \cellcolor{mygray}56.0          & \cellcolor{mygray}49.6          & \cellcolor{mygray}48.7          & \cellcolor{mygray}46.8          \\
                                   & POCON \cite{gomez2024plasticity}                                & \textbf{63.7}                         & 60.5                                  & 56.8                                  & 48.9                                  & 48.9                                  \\
\multirow{-6}{*}{\begin{tabular}[c]{@{}c@{}}CIFAR\\ 100\end{tabular}}        & \cellcolor{mygray}POCON +BT      & \cellcolor{mygray}63.1 & \cellcolor{mygray}\textbf{61.1} & \cellcolor{mygray}\textbf{58.7} & \cellcolor{mygray}\textbf{51.6} & \cellcolor{mygray}\textbf{51.4} \\ \hline
                                   & Fine-tuning                           & 42.0                                  & 36.6                                  & 32.3                                  & 22.3                                  & 2.8                                   \\
                                   & \cellcolor{mygray}Branch-tuning & \cellcolor{mygray}44.4          & \cellcolor{mygray}38.1          & \cellcolor{mygray}36.8          & \cellcolor{mygray}26.1          & \cellcolor{mygray}22.9          \\
                                   & PFR \cite{gomez2022continually}                                  & 42.2                                  & 39.2                                  & 31.2                                  & 25.9                                  & 21.2                                  \\
                                   & \cellcolor{mygray}PFR +BT        & \cellcolor{mygray}42.7          & \cellcolor{mygray}38.5          & \cellcolor{mygray}37.7          & \cellcolor{mygray}28.3          & \cellcolor{mygray}23.1          \\
                                   & POCON \cite{gomez2024plasticity}                                & 41.0                                  & 41.1                                  & 41.1                                  & 37.2                                  & 30.2                                  \\
\multirow{-6}{*}{\begin{tabular}[c]{@{}c@{}}Tiny\\ ImageNet\end{tabular}}     & \cellcolor{mygray}POCON +BT      & \cellcolor{mygray}\textbf{45.8} & \cellcolor{mygray}\textbf{42.8} & \cellcolor{mygray}\textbf{42.5} & \cellcolor{mygray}\textbf{38.5} & \cellcolor{mygray}\textbf{33.9} \\ \hline
                                   & Fine-tuning                           & 56.1                                  & 48.1                                  & 42.7                                  & 39.6                                  & 21.0                                  \\
                                   & \cellcolor{mygray}Branch-tuning & \cellcolor{mygray}65.4          & \cellcolor{mygray}55.5          & \cellcolor{mygray}50.3          & \cellcolor{mygray}44.7          & \cellcolor{mygray}37.4          \\
                                   & PFR \cite{gomez2022continually}                                  & 66.1                                  & 60.5                                  & 54.8                                  & 42.2                                  & 38.3                                  \\
                                   & \cellcolor{mygray}PFR +BT        & \cellcolor{mygray}66.3          & \cellcolor{mygray}61.3          & \cellcolor{mygray}56.5          & \cellcolor{mygray}44.1          & \cellcolor{mygray}42.0          \\
                                   & POCON \cite{gomez2024plasticity}                                & \textbf{66.3}                         & 61.4                                  & 59.3                                  & 53.5                                  & 45.4                                  \\
\multirow{-6}{*}{\begin{tabular}[c]{@{}c@{}}ImageNet\\ 100\end{tabular}}     & \cellcolor{mygray}POCON +BT      & \cellcolor{mygray}66.1          & \cellcolor{mygray}\textbf{62.6} & \cellcolor{mygray}\textbf{61.2} & \cellcolor{mygray}\textbf{55.3} & \cellcolor{mygray}\textbf{48.6} \\ 
\bottomrule[1.3pt]
\end{tabular}
\vskip -0.1in
\end{table}

\begin{table}[htb]

   \caption{Linear-Probe accuracy on CIFAR-100 and ImageNet-100 after 5-task data-incremental SSL.}
    \label{table:3}
    \vskip -0.1in
 \small \centering 
 \renewcommand\tabcolsep{6pt}
 \renewcommand{\arraystretch}{1.3}
\begin{tabular}{clcc}
\toprule[1.3pt] 
\textbf{Method}                                                          & \textbf{Strategy}                     & \textbf{CIFAR-100}                     & \textbf{ImageNet-100}                 \\ \hline
                                                                         & Fine-tuning                           & 54.5                                   & 68.9                                  \\
                                                                         & \cellcolor{mygray}Branch-tuning & \cellcolor{mygray}\textbf{57.0} & \cellcolor{mygray}\textbf{71.6} \\
\multirow{-3}{*}{SimCLR}                                                 & Offline                               & 65.8                                   & 77.5                                  \\ \hline
                                                                         & Fine-tuning                           & 58.3                                   & 74.0                                    \\
                                                                         & \cellcolor{mygray}Branch-tuning & \cellcolor{mygray}\textbf{60.8} & \cellcolor{mygray}\textbf{76.1} \\
\multirow{-3}{*}{BYOL}                                                   & Offline                               & 70.5                                   & 80.3                                  \\ \hline
                                                                         & Fine-tuning                           & 57.7                                   & 71.3                                  \\
                                                                         & \cellcolor{mygray}Branch-tuning & \cellcolor{mygray}\textbf{60.8} & \cellcolor{mygray}\textbf{73.3} \\
\multirow{-3}{*}{\begin{tabular}[c]{@{}c@{}}Barlow\\ Twins\end{tabular}} & Offline                               & 70.9                                   & 80.4                                  \\ \hline
                                                                         & Fine-tuning                           & 53.6                                   & 69.5                                  \\
                                                                         & \cellcolor{mygray}Branch-tuning & \cellcolor{mygray}\textbf{57.0} & \cellcolor{mygray}\textbf{71.8} \\
\multirow{-3}{*}{MoCoV2+}                                                & Offline                               & 69.9                                   & 77.5                                  \\ 
  \bottomrule[1.3pt]
\end{tabular}
\vskip -0.1in
\end{table}

\begin{table}[htb]

   \caption{Linear-Probe accuracy of transfer on RAF-DB real-world dataset, evaluated at each stage of the 5-task class-incremental SSL on CIFAR-100.}
    \label{table:4}
    \vskip -0.1in
 \small \centering 
 \renewcommand\tabcolsep{2.8pt}
 \renewcommand{\arraystretch}{1.3}
\begin{tabular}{clccccc}
\toprule[1.3pt] 
\textbf{Method}           & \textbf{Strategy}                     & \textbf{stage1}                        & \textbf{stage2}                        & \textbf{stage3}                        & \textbf{stage4}                        & \textbf{stage5}                        \\ \hline
                          & Fine-tuning                           & 53.4                                  & 53.2                                  & 53.5                                  & 53.0                                  & 54.1                                  \\
                          & \cellcolor{mygray}Branch-tuning & \cellcolor{mygray}\textbf{53.4} & \cellcolor{mygray}\textbf{55.3} & \cellcolor{mygray}\textbf{56.1} & \cellcolor{mygray}\textbf{56.5} & \cellcolor{mygray}\textbf{57.1} \\
\multirow{-3}{*}{SimCLR}  & Offline                               &                                       &                                       &                                       &                                       & 57.9                                  \\ \hline
                          & Fine-tuning                           & 59.1                                  & 57.6                                  & 55.4                                  & 55.8                                  & 57.6                                  \\
                          & \cellcolor{mygray}Branch-tuning & \cellcolor{mygray}\textbf{59.1} & \cellcolor{mygray}\textbf{60.2} & \cellcolor{mygray}\textbf{60.9} & \cellcolor{mygray}\textbf{61.6} & \cellcolor{mygray}\textbf{61.9} \\
\multirow{-3}{*}{BYOL}    & Offline                               &                                       &                                       &                                       &                                       & 62.0                         \\ \hline
                          & Fine-tuning                           & 56.3                                  & 57.3                                  & 55.1                                  & 55.9                                  & 55.2                                  \\
                          & \cellcolor{mygray}Branch-tuning & \cellcolor{mygray}\textbf{56.3} & \cellcolor{mygray}\textbf{58.3} & \cellcolor{mygray}\textbf{58.7} & \cellcolor{mygray}\textbf{58.6} & \cellcolor{mygray}\textbf{59.1} \\
\multirow{-3}{*}{Barlow}  & Offline                               & \textbf{}                             & \textbf{}                             & \textbf{}                             & \textbf{}                             & 60.1                         \\ \hline
                          & Fine-tuning                           & 56.0                                  & 53.8                                  & 53.9                                  & 52.8                                  & 54.0                                  \\
                          & \cellcolor{mygray}Branch-tuning & \cellcolor{mygray}\textbf{56.0} & \cellcolor{mygray}\textbf{56.7} & \cellcolor{mygray}\textbf{57.2} & \cellcolor{mygray}\textbf{57.5} & \cellcolor{mygray}\textbf{58.2} \\
\multirow{-3}{*}{MoCoV2+} & Offline                               &                                       &                                       &                                       &                                       & 59.3                                  \\ \hline
Random                    &                                       &                                       &                                       &                                       &                                       & 44.8                                 \\ 
\bottomrule[1.3pt]
\end{tabular}
    \vskip -0.1in
\end{table}

\vspace*{5pt}

\begin{figure*}[t]
  \centering
  \includegraphics[width=1.0\linewidth]{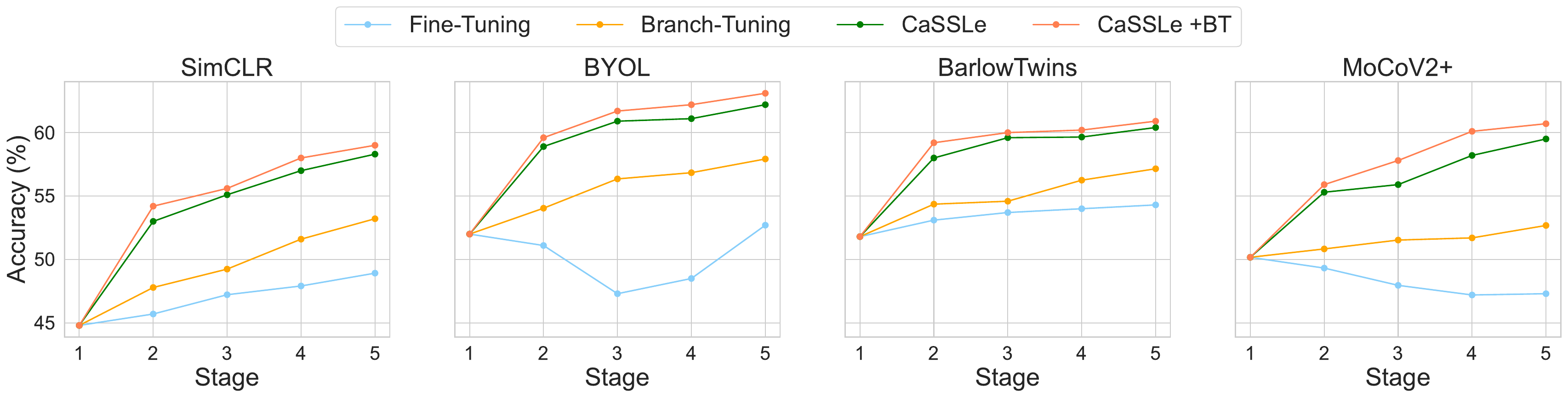}
  \vskip -0.1in
  \caption{Linear-Probe evaluation on CIFAR-100, which involves 5 tasks in a class-incremental setting. Compared with baseline Fine-tuning and state-of-the-art method CaSSLe \cite{fini2022self}.}
  \label{fig:7}
\end{figure*}

\noindent\textbf{Comparison with CaSSLe.} We compared Branch-tuning (BT) with CaSSLe under the class-incremental setting, which simulates scenarios where new categories emerge gradually. As shown in Table \ref{table:3} and Figure \ref{fig:7}, without retaining the old model and data, nor employing knowledge distillation, BT significantly outperforms Fine-tuning (FT), achieving accuracy improvements of up to 5.4\% and 3.9\% on CIFAR-100 and ImageNet-100, respectively. 
By preserving the old model and implementing knowledge distillation during training on new data, the state-of-the-art method CaSSLe shows notable improvements on the baseline performance, with increases of 12.2\% on CIFAR-100 and 6.5\% on ImageNet-100.
CaSSLe can improve its performance easily by adopting Branch-tuning, which can result in performance improvements ranging from 1.1\% to 1.2\%. 
Similar enhancements are observed in other metrics, such as accuracy and forgetting rate. This demonstrates that our method is adaptable to different settings, achieving state-of-the-art performance with FT or CaSSLe models.


\vspace*{5pt}
\noindent\textbf{Comparison with other continual SSL methods.}
In the class-incremental setting of POCON, we compared the performance of BT, FT, PFR \cite{gomez2022continually}, and POCON \cite{gomez2024plasticity}. Notably, POCON comprises three learning phases, wherein we applied BT during the first phase (Self-Supervised Learning in training data). As shown in Table \ref{table:pocon}, BT demonstrated significant performance improvements across more task splits and categories, such as TinyImageNet. Compared to the baseline method FT, BT achieved an improvement of up to 20.1\% on TinyImageNet.
BT training does not require saving old models or data, nor knowledge distillation. It even surpasses PFR by 1.7\%.
When applying BT to the latest methods PFR and POCON, the performance of POCON +BT is slightly lower than POCON by 0.2-0.6\% in the four-task setting. However, BT showcases marked improvements in class incremental setting of 10 to 100 tasks, with gains of up to 2.5\% on CIFAR-100, 3.7\% on TinyImageNet, and 3.2\% on ImageNet.
This indicates that BT offers substantial performance boosts in scenarios with growing numbers of classes and tasks, and can be directly applied to complex models based on knowledge distillation. 

\vspace*{5pt}
\noindent\textbf{Comparison in data-incremental setting.} In the data-incremental setting, the gap between continuous learning and joint training for SSL models narrows, as access to all-class data in each task reduces catastrophic forgetting. In this context, as illustrated in Table \ref{table:3}, Branch-tuning achieves promising results with accuracy improvements of up to 3.1\% and 2.3\% on CIFAR-100 and ImageNet-100, respectively. This further reduces the discrepancy between continual SSL and retraining on all data.

\vspace*{5pt}
\noindent\textbf{Comparison on real-world dataset.} To validate the effectiveness of Branch-tuning in real-world applications, we conduct transfer learning on the RAF-DB dataset for each stage of continual SSL. Compared to randomly initialized models, BYOL attains the highest improvement in linear-probe accuracy, with an increase of up to 17.2\% through unsupervised pretraining. As shown in Table \ref{table:4}, our method achieves improvements of 3.0\%, 4.3\%, 3.9\%, and 4.2\% for SimCLR, BYOL, BarlowTwins, and MoCoV2+ in continual SSL pretraining, respectively. With Branch-tuning, the performance gap between continual SSL and offline training is notably reduced.

\subsection{Ablation Studies}

In this section, we analyze ablation studies on Branch-tuning structures and fixed BN layers' impact, using the CIFAR-100 dataset.


\vspace*{5pt}
\noindent\textbf{Branch-tuning Structures:} To examine the effect of branch structure on Branch-tuning, we performed incremental SSL experiments over five stages using three different structures: 1x1, 1x3, and 3x3. Intuitively, a larger parameter size in the branch structure leads to increased plasticity and reduced stability, and vice versa. This conclusion is supported by Figure \ref{fig:2} in Section \ref{sec:4}. In most SSL methods, the 1x3 structure achieves a superior balance and higher linear-probe accuracy.

\begin{figure}[t]
  \centering
  \includegraphics[width=1.0\linewidth]{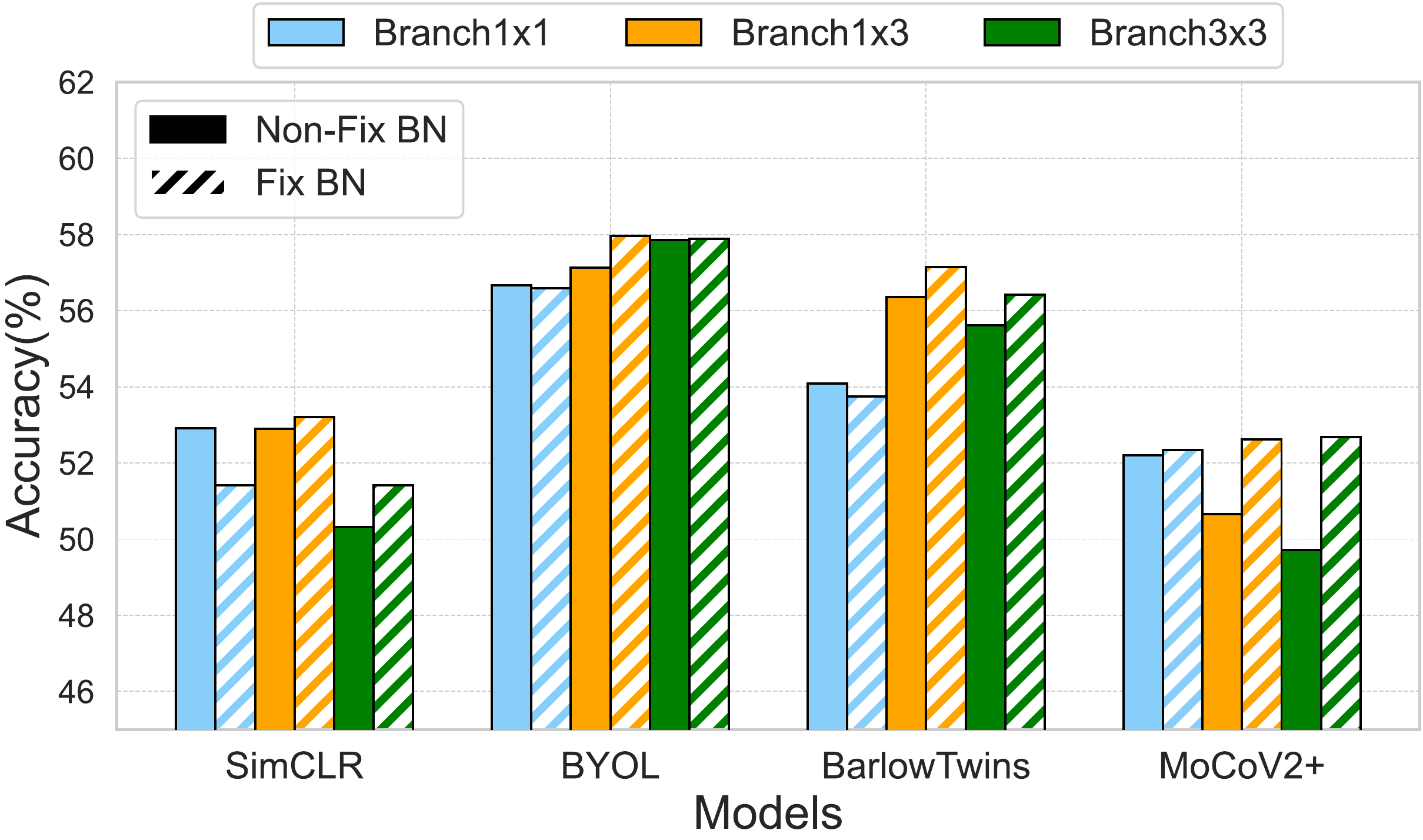}
  \vskip -0.1in
  \caption{Linear-Probe evaluation of different settings for BN Layers and branch structures on CIFAR-100.}
  \vskip -0.1in
  \label{fig:8}
\end{figure}

\vspace*{5pt}
\noindent\textbf{Effect of Fixing BN layers:} We investigate the impact of fixing the BN layers in Branch-tuning during continual SSL. As shown in Table \ref{table:5} and Figure \ref{fig:8}, for 1x3 and 3x3 branch structures, fixing BN layers has a positive effect, leading to improvements of up to 3.0\%. However, for the 1x1 branch structure, fixing batch normalization has a negative impact. The 1x1 branch structure inherently possesses strong stability. Fixing batch normalization further strengthens it, affecting the balance between plasticity and stability in the continual SSL process and resulting in an overall decline in linear-probe accuracy.

\begin{table}[tb]

   \vskip -0.1in
   \caption{Linear-Probe accuracy on CIFAR-100 after 5-task data-incremental SSL, comparing fixed vs. non-fixed BN and different branch structures.}
    \label{table:5}
       \vskip -0.1in
 \small \centering 
 \renewcommand\tabcolsep{6pt}
 \renewcommand{\arraystretch}{1.3}
\begin{tabular}{ccccccc}
\toprule[1.3pt] 
\multirow{2}{*}{\textbf{\begin{tabular}[c]{@{}c@{}}Branch-tuning\\ Setting\end{tabular}}} & \multicolumn{3}{c}{\textbf{Non-Fix BN}} & \multicolumn{3}{c}{\textbf{Fix BN}}     \\ \cline{2-7} 
                                                                                          & 1x1           & 1x3           & 3x3          & 1x1   & 1x3            & 3x3            \\ \hline
SimCLR                                                                                    & 52.9         & 52.9         & 50.3        & 51.4 & \textbf{53.2} & 51.4 \\
BYOL                                                                                      & 56.7         & 57.1         & 56.8        & 56.6 & \textbf{58.0} & 57.9          \\
Barlow                                                                                    & 54.1         & 56.4         & 55.6        & 53.8 & \textbf{57.2} & 56.4          \\
MoCoV2+                                                                                   & 52.2          & 50.7         & 49.7        & 52.3 & 52.6          & \textbf{52.7} \\ 
\bottomrule[1.3pt]
\end{tabular}
\end{table}

\subsection{Further Analysis and Discussion}

\begin{table*}[htb]
   \caption{Semi-supervised Linear-Probe accuracy on ImageNet-100 after 5-task class-incremental SSL.}
    \label{table:7}
       \vskip -0.1in
 \small \centering 
 \renewcommand\tabcolsep{6pt}
 \renewcommand{\arraystretch}{1.3}
\begin{tabular}{clccccc}
\toprule[1.3pt] 
\textbf{Percentage}    & \textbf{Strategy}                     & \textbf{SimCLR}                       & \textbf{BYOL}                         & \textbf{Barlow Twins}                      & \textbf{MoCoV2+}                      & \textbf{Supervised}    \\ \hline
                       & Fine-tuning                           & 39.7                                  & 42.3                                  & 42.6                                  & 40.9                                  &                        \\
\multirow{-2}{*}{1\%}  & \cellcolor{mygray}Branch-tuning & \cellcolor{mygray}\textbf{42.8} & \cellcolor{mygray}\textbf{43.1} & \cellcolor{mygray}\textbf{45.9} & \cellcolor{mygray}\textbf{44.5} & \multirow{-2}{*}{48.1} \\ 
                       & Fine-tuning                           & 49.4                                  & 51.2                                  & 51.9                                  & 50.2                                  &                        \\
\multirow{-2}{*}{5\%}  & \cellcolor{mygray}Branch-tuning & \cellcolor{mygray}\textbf{51.7} & \cellcolor{mygray}\textbf{52.4} & \cellcolor{mygray}\textbf{53.2} & \cellcolor{mygray}\textbf{51.8} & \multirow{-2}{*}{55.6} \\ 
                       & Fine-tuning                           & 52.5                                  & 55.7                                  & 56.6                                  & 54.9                                  &                        \\
\multirow{-2}{*}{10\%} & \cellcolor{mygray}Branch-tuning & \cellcolor{mygray}\textbf{54.7} & \cellcolor{mygray}\textbf{56.2} & \cellcolor{mygray}\textbf{58.4} & \cellcolor{mygray}\textbf{58.6} & \multirow{-2}{*}{60.8} \\ 
\bottomrule[1.3pt]
\end{tabular}
  \vskip -0.1in
\end{table*}

\vspace*{5pt}
\noindent\textbf{Additional results with semi-supervised.}
We conducted an evaluation of continual SSL models' semi-supervised performance. Our evaluation focused on models obtained from the final stage of the ImageNet-100 five-stage class increment setup. These models were linear evaluate on a small number of labeled samples from ImageNet-100 to simulate real-world scenarios with insufficient sample labels.
As shown in Table \ref{table:7},  Branch-tuning resulted in better performance across various semi-supervised settings. Even in extremely scarce labeled samples, such as when only 1\% of samples were labeled, Branch-tuning improved model accuracy by 0.8-3.6\%. The model’s performance significantly improved when we increased the number of labeled samples to 5\%. Under this setting, the model's accuracy increased by up to 8.9-9.7\%, and our method achieved a 1.2-2.3\% enhancement. At 10\% labeled samples, the model's performance improved further, with an accuracy up to 58.6\%, and Branch-tuning boosted semi-supervised performance by 0.5-3.7\%. 

\begin{figure}[t]
  \centering
  \includegraphics[width=0.9\linewidth]{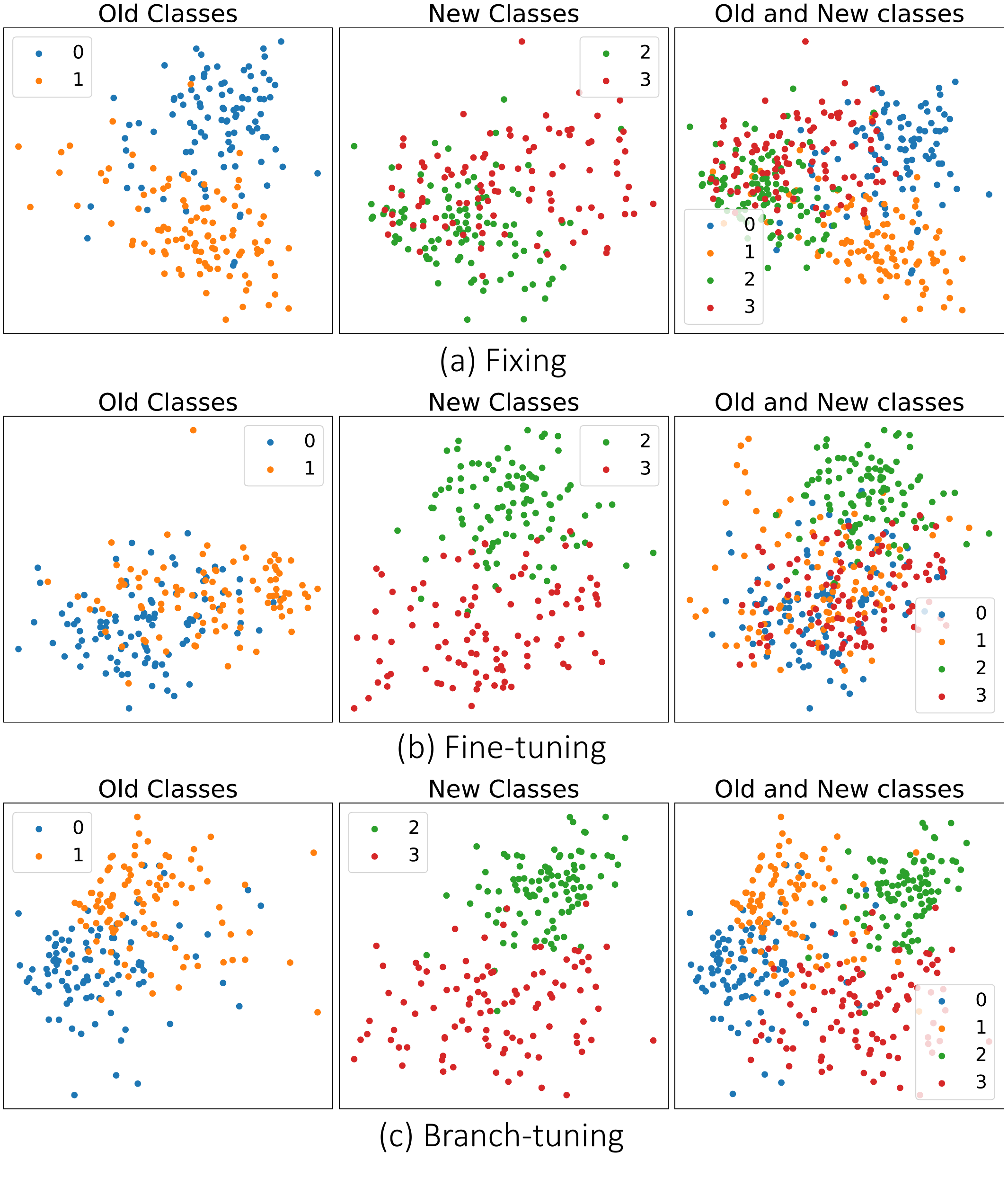}
  \vskip -0.1in
  \caption{Using PCA to visualize the data of four categories in CIFAR-100, it can be seen that the features extracted by Fine-tuning or Fixing the model result in more significant feature confusion.}
  \vskip -0.2in
  \label{fig:9}
\end{figure}

%
%
%
%
%

\begin{table}[t]
    \begin{minipage}{.5\linewidth}
        \caption{Training time comparison of repeat train on 5-task data flow and continual SSL.}
        \label{table:8}
        \centering
        \small 
        \renewcommand\tabcolsep{2pt}
        \renewcommand{\arraystretch}{1.3}
        \begin{tabular}{lc}
            \toprule[1.3pt] 
            \textbf{Strategy} & \textbf{Time}  \\ \hline
            Repeat Train      & $15\times T_0$       \\
            Continual SSL     & \textbf{$5\times T_0$} \\ 
            \bottomrule[1.3pt]
        \end{tabular}
    \end{minipage}%
    \begin{minipage}{.5\linewidth}
        \caption{Training time comparison of Branch-tuning and Fine-tuning.}
        \label{table:9} 
        \centering
        \small 
        \renewcommand\tabcolsep{2pt}
        \renewcommand{\arraystretch}{1.3}
        \begin{tabular}{lcc}
            \toprule[1.3pt] 
            \textbf{Strategy} & \textbf{Params } & \textbf{$T_0$ } \\ \hline
            Fine-tuning       & 11.1M           & 135.1s \\
            Branch-tuning     & \textbf{7.8M}   & \textbf{113.7s} \\ 
            \bottomrule[1.3pt]
        \end{tabular}
    \end{minipage}
      \vskip -0.1in
\end{table}

\vspace*{5pt}
\noindent\textbf{Visualization of feature space.}
We conducted a PCA visualization to compare the models in the first task ($t_0$) and the last task ($t_4$) of a five-stage class incremental scenario on CIFAR-100. After analyzing the Branch-tuning, Fine-tuning, and Fixed models, we observed some key points in Figure \ref{fig:9}. 
Firstly, Fine-tuning caused more confusion on the old category data than Branch-tuning. This could be because Fine-tuning might interfere with the model's understanding of the original tasks, thus increasing confusion in the feature space of the old categories. 
On the other hand, the Fixed model created confusion in the feature space of the new categories. This could be due to the inability to adjust the model's parameters when facing new tasks, making it ineffective in handling the new category data and causing confusion in the feature space of the new classes. 
In contrast, Branch-tuning demonstrated better differentiation abilities on both old and new categories. This indicates that Branch-tuning achieved a good balance between maintaining model stability and enhancing plasticity.

\vspace*{5pt}
\noindent\textbf{Training overhead.}
SSL requires more resources, data, and training epochs than supervised learning. This means that the time it takes to train a model should be considered. We focus on two key issues: \textbf{(1)} the time it takes to continual SSL versus repeated data collection training, and \textbf{(2)} comparing the training time between Branch-tuning and Fine-tuning.
To examine this, we conducted 5 task training using the CIFAR-100 dataset. We assigned a basic unit of training time, denoted as $T_0$, for each task. Our findings show that when the model is repeatedly trained on a continuous data stream, the training time increases by 3 times.
Branch-tuning can reduce training time overhead by reducing the number of gradients to be recorded and parameters updated. As shown in Table \ref{table:9}, it reduces training parameters by 29.7\% and training time by 15.8\% for unit time $T_0$ using ResNet18 and SimCLR models.

\section{Conclusion}
\label{sec:7}
In real-world scenarios, non-IID and infinite data constantly emerge, necessitating SSL models to continuously learn rather than repetitive training. To address the model's stability-plasticity dilemma when learning from new data, we utilize CKA to quantify these aspects, finding that BN layers are crucial for stability, while convolutional layers play a vital role in plasticity. Based on these insights, we propose Branch-tuning, which includes branch expansion and compression. Branch expansion enables the model to learn new knowledge without impacting old parameters, while branch compression allows the branch to be equivalently integrated into the original network structure, preserving its consistency. Our method effectively balances stability and plasticity in continual SSL without requiring storage for old data or models and is easy to integrate into existing SSL and continual SSL methods.

Future research could explore techniques to optimize the stability-plasticity balance, assess Branch-tuning's applicability to various neural network architectures, and investigate potential benefits of combining Branch-tuning with other continual learning strategies to advance continual SSL methods.

\bibliography{refer}
\bibliographystyle{IEEEtran}


%
%
%
%

\vfill

\end{document}